\documentclass[graybox]{svmult}

\usepackage{mathptmx}       
\usepackage{helvet}         
\usepackage{courier}        
\usepackage{type1cm}        
\usepackage{makeidx}         
\usepackage{graphicx}        
\usepackage{multicol}        
\usepackage[bottom]{footmisc}

\usepackage{csquotes}
\usepackage[T1]{fontenc}
\usepackage[scaled=0.83]{beramono}
\usepackage {listings,soul}
\usepackage{todonotes}
\usepackage{algorithm}%
\usepackage{algorithmicx}%
\usepackage{algpseudocode}%
\usepackage{amsmath,amssymb,amsfonts}%
\usepackage{booktabs}%

\newcommand{\code}[1]{\texttt{#1}}
\newcommand{\evolv}[1]{\textcolor{olive}{\underline{#1}}}

\makeindex             

\begin{document}

\title*{Origami: (un)folding the abstraction of recursion schemes for program synthesis}
\titlerunning{Origami: (un)folding the abstraction of recursion schemes for program synthesis}
\author{Matheus Campos Fernandes, Fabricio Olivetti de Franca, Emilio Francesquini}
\institute{
Matheus Campos Fernandes \at Federal University of ABC, Santo Andre, SP, Brazil, \email{fernandes.matheus@ufabc.edu.br}
\and Fabricio Olivetti de Franca \at Federal University of ABC, Santo Andre, SP, Brazil \email{folivetti@ufabc.edu.br}
\and Emilio Francesquini \at Federal University of ABC, Santo Andre, SP, Brazil  \email{e.francesquini@ufabc.edu.br}
}
%
%
\maketitle

\abstract*{Program synthesis with Genetic Programming searches for a correct program that satisfies the input specification, which is usually provided as input-output examples. One particular challenge is how to effectively handle loops and recursion avoiding programs that never terminate.
A helpful abstraction that can alleviate this problem is the employment of Recursion Schemes that generalize the combination of data production and consumption.
Recursion Schemes are very powerful as they allow the construction of programs that can summarize data, create sequences, and perform advanced calculations.
The main advantage of writing a program using Recursion Schemes is that the programs are composed of well defined templates with only a few parts that need to be synthesized.
In this paper we make an initial study of the benefits of using program synthesis with \emph{fold} and \emph{unfold} templates, and outline some preliminary experimental results. To highlight the advantages and disadvantages of this approach, we manually solved the entire GPSB benchmark using recursion schemes, highlighting the parts that should be evolved compared to alternative implementations.
We noticed that, once the choice of which recursion scheme is made, the synthesis process can be simplified as each of the missing parts of the template are reduced to simpler functions, which are further constrained by their own input and output types.}

\abstract{Program synthesis with Genetic Programming searches for a correct program that satisfies the input specification, which is usually provided as input-output examples. One particular challenge is how to effectively handle loops and recursion avoiding programs that never terminate.
A helpful abstraction that can alleviate this problem is the employment of Recursion Schemes that generalize the combination of data production and consumption.
Recursion Schemes are very powerful as they allow the construction of programs that can summarize data, create sequences, and perform advanced calculations.
The main advantage of writing a program using Recursion Schemes is that the programs are composed of well defined templates with only a few parts that need to be synthesized.
In this paper we make an initial study of the benefits of using program synthesis with \emph{fold} and \emph{unfold} templates, and outline some preliminary experimental results. To highlight the advantages and disadvantages of this approach, we manually solved the entire GPSB benchmark using recursion schemes, highlighting the parts that should be evolved compared to alternative implementations.
We noticed that, once the choice of which recursion scheme is made, the synthesis process can be simplified as each of the missing parts of the template are reduced to simpler functions, which are further constrained by their own input and output types.}

\section{Introduction}\label{sec1}

Computer programming can be seen as the task of creating a set of instructions that, when executed, can provide as output a solution to a specific problem. This task involves several steps starting from an abstract description of the solution (\emph{i.e.}, an algorithm) to a concrete implementation written in a programming language.

Given the importance of creating computer programs, and the repetitive tasks usually involved, an often sought \emph{holy grail} is the ability to automatically generate source-codes, either in part or in its entirety. This automatic generation would follow a certain high-level specification which reduces the burden of manually creating the program. This problem is known as \emph{Program Synthesis} (PS). In the specific scenario in which the user provides a set of input-output examples, the problem is referred to as \emph{Inductive Synthesis} or \emph{Programming-by-Example}. The main advantage of this approach is that it can be easier to create a set of inputs and expected outputs but, on the other hand, it might be difficult to provide a representative set containing corner cases.
A popular approach for PS is Genetic Programming (GP), a technique that applies the concept of evolution to search for a program constrained by the search space of a solution representation. Some notable approaches include PushGP~\cite{helmuth2018program}, CBGP~\cite{pantridge2020code}, GE~\cite{o2001grammatical}, and GGGP~\cite{manrique2009grammar}.

So far, most of the PS algorithms do not effectively exploit common programming patterns, which means the search algorithm needs to scan the search space to find a correct program without any guidance. While a less restrictive search space can be desirable to allow the algorithm to navigate towards one of the many solutions, a constrained search space, if correctly done, can speed up the search process allowing the search algorithm to focus only on part of the program.
Fernandes et al.~\cite{fernandes2023hotgp} explored the use of abstractions (\emph{e.g.}, higher-order functions) and additional information (\emph{e.g.}, type information) to help with the exploration of the search space with the introduction of the Higher-Order Typed Genetic Programming  (HOTGP). This algorithm exploits the information of the input and output types of the desired program to limit the search space. It also adds support to higher-order functions, $\lambda$-functions, and parametric polymorphism to make it possible to apply common programming patterns such as \emph{map} and \emph{filter}. The authors showed that the use of the type information and higher-order functions helped to improve the success rate on GPSB, surpassing some of the current state-of-the-art approaches.

A general and useful pattern is the use of Recursion Schemes~\cite{meijer1991functional}. This pattern captures the common structure of recursive functions and data structures as combinations of \emph{consumer} and \emph{producer} functions, also known as \emph{fold} and \emph{unfold}, respectively. They are known to be very general and capable of implementing many commonly used algorithms, ranging from data aggregation to sorting. Gibbons~\cite{gibbons2003fun} coined the name \emph{origami programming} and showed many examples of how to write common algorithms using these patterns.
The folding and unfolding process can be generalized through Recursion Schemes, which divide the programming task into three simpler steps: (\emph{i}) choosing the scheme (among a limited number of choices), (\emph{ii}) choosing a fixed point of a data structure that describes the recursion trace and, (\emph{iii}) writing the consumer (fold) and producer (unfold) procedure.

In~\cite{swan2019stochastic}, the authors introduced the idea of taking advantage of recursion schemes to constrain the search space for program synthesis. In their work, they assumed a prior knowledge, using a human in the loop, to correctly choose the algebraic data type and the recursion scheme. Then the program synthesizer would automatically infer the grammar from the choice of type and recursion scheme, as they are uniquely determined, and then search for the correct cases heuristically. They tested the idea for some variations of Fibonacci algorithm and some programs applied to the Nat ADT together with the catamorphism, showing a higher hit rate than other approaches.

In this paper we studied the problems in the General Program Synthesis Benchmark Suite (GPSB)~\cite{helmuth2015general} and solved them using Recursion Schemes, reporting a selection of distinct solutions. This examination revealed that most of the solutions for the proposed problems follow the common pattern of folding, unfolding a data structure, or a composition of both.
With these observations, we explore the crafting of computer programs following these recursive patterns and craft program templates that follow these patterns along with an explanation of how they can constrain the search space of candidate programs. Our main goal is to find a set of recursion schemes that simplifies the program synthesis process while reducing the search space with type information. We also present the general idea on how to evolve such programs, called Origami, a program synthesis algorithm that first determines the (un)folding pattern it will evolve and then it evolves the corresponding template using inductive synthesis.

The remainder of this text is organized as follows. In Section~\ref{sec:recschemes} we introduce recursion schemes and explain the basic concepts needed to understand our proposal. Section~\ref{sec:origami} presents Origami and outlines some examples of recursive patterns that can be used to solve common programming problems. In Section~\ref{sec:preliminary} we show some preliminary results adapting HOTGP to one of the presented patterns and analyze the results. Finally, in Section~\ref{sec:discussion} we give some final observations about Origami and describe future work.

\section{Recursion Schemes}\label{sec:recschemes}

Recursive functions are sometimes referred to as the \emph{goto} of functional programming languages~\cite{jones2001expressive,harvey1992avoiding,meijer1991functional,gibbons2003fun}. They are an essential part of how to build a program in these languages, allowing programmers to avoid mutable state and other imperative programming constructs.
Both \emph{goto} and recursion are sometimes considered harmful when creating a computer program. The reason is that the former can make it hard to understand the program execution flow as the program grows larger, and the latter may lead to problems such as \emph{stack overflows} if proper care is not taken. This motivated the introduction of higher-order functions as the preferred alternative to direct recursion.

Higher-order functions are functions that either (\emph{i}) take one or more functions as an argument, or (\emph{ii}) return a function as output. Here we are interested mostly in the first case. By using higher-order functions which apply the function received as input in a recursive pattern to a given value or data structure, we can represent many common recursion schemes. Some of the most well-known examples of these functions are \emph{map}, \emph{filter}, and \emph{fold}. The main advantage of using these general patterns is that the programmer does not need to be as careful (guaranteeing termination, or sane memory usage for example) as they would need to be using direct recursion.

Among these higher-order functions, \emph{fold} is the most general (as in it can be used to implement the others). This pattern is capable of representing many recursive algorithms that start with a list of values and returns a transformed value. Another common general pattern is captured by the \emph{unfold} function that starts from a seed value and unfolds it generating a list of values in the process.

While \emph{fold} and \emph{unfold} describe most common patterns of recursive functions, they are limited to recursions that follow a linear path (\emph{e.g.}, a list). In some scenarios, the recursion follows a nonlinear path such as a binary tree (\emph{e.g.}, most comparison-based sort algorithms). To generalize the \emph{fold} and \emph{unfold} operations to different recursive patterns, Recursion Schemes~\cite{garland1973program,meijer1991functional} describe common patterns of generating and consuming inductive data types, not limited to only the consumption or generation of data, but also abstracting the idea of having access to partial results or even backtracking. The main idea is that the recursion path is described by a fixed point of an inductive data type and the program building task becomes limited to some specific definitions induced by the chosen data structure. This concept is explained in more detail, in the following section.

\subsection{Fixed point of a linked list}

A generic inductive list (\emph{i.e.}, linked list) carrying values of type \code{a} is described as\footnote{In this paper we employ the Haskell language notation, which is similar to the ML notation.}:

\begin{lstlisting}
    data List a = Nil | Cons a (List a)
\end{lstlisting}

This is read as a list can be either empty (\code{Nil}) or a combination (using \code{Cons}) of a value of type \code{a} followed by another list (\emph{i.e.}, its tail). In this context, \code{a} is called a type parameter\footnote{Similar to generics in Java and templates in C++.}. We can eliminate the recursive definition by adding a second parameter replacing the recursion:

\begin{lstlisting}
    data ListF a b = NilF | ConsF a b
\end{lstlisting}

This definition still allows us to carry the same information as \code{List} but, with the removal of the recursion from the type, it becomes impossible to create generic functions that work with different list lengths. Fortunately, we can solve this problem by obtaining the fixed point of \code{ListF}, that corresponds to the definition of \code{List}:

\begin{lstlisting}
    data Fix f = MkFix (f (Fix f))

    unfix :: Fix f -> f (Fix f)
    unfix (MkFix x) = x
\end{lstlisting}

\code{MkFix} is a \emph{type constructor} and \code{unfix} extracts one layer of our nested structure. 
The \code{Fix} data type creates a link between the fixed point of a list and the list itself. This allows us to write recursive programs for data structures with non-recursive definitions that are very similar to those targeting data structures with recursive definitions.

\subsection{Functor Algebra}

Recursion schemes come from two basic operations: the \emph{algebra}, that describes how elements must be consumed; and \emph{coalgebra}, describing how elements must be generated.
Assuming the fixed point structure \code{f a} supports a function \code{fmap}, a higher-order function that applies the first argument to every nested element of the structure \code{f}\footnote{In category theory this is called an \emph{endofunctor}.} a Functor Algebra (F-Algebra) and its dual, Functor Co-Algebra, are defined as:

\begin{lstlisting}
    data Algebra f a = f a -> a
    data CoAlgebra f a = a -> f a
\end{lstlisting}

Or, in plain English, it is a function that combines all the information carried by the parametric type (\code{f}) into a single value (\emph{i.e.}, \code{fold} function) and a function that, given a seed value, creates a data structure defined by \code{f} (i.e., \code{unfold} function).
The application of an Algebra into a fixed point structure is called \emph{catamorphism} and \emph{anamorphism} for Co-algebras:

\begin{lstlisting}
    cata :: (f a -> a) -> Fix f -> a
    cata alg data = alg (fmap (cata alg) (unFix data))

    ana :: (a -> f a) -> a -> Fix f
    ana coalg seed = MkFix (fmap (ana coalg) (coalg seed))

    -- definition of fmap for ListF
    fmap :: (b -> c) -> ListF a b -> ListF a c
    fmap f NilF = NilF
    fmap f (ConsF a b) = ConsF a (f b)
\end{lstlisting}

 So, given an algebra \code{alg}, \code{cata} \emph{peels} the outer layer of the fixed-point \code{data}, maps itself to the whole structure, and applies the algebra to the result. In short, the procedure traverses the structure to its deepest layer and applies \code{alg} recursively accumulating the result. Similarly, \code{ana} can be seen as the inverse of \code{cata}. In this function, we first apply \code{coalg} to \code{seed}, generating a structure of type \code{f} (usually a singleton), then we map the \code{ana coalg} function to the just generated data to further expand it, finally we enclose it inside a \code{Fix} structure to obtain the fixed point. In the context of lists, this procedure is known as \code{unfold} as it departs from one value and unrolls it into a list of values.
 Although the general idea of defining the fixed-point of a data structure and implementing the catamorphism may look like over-complicating standard functions, the end result allows us to focus on much simpler implementations. In the special case of a list, we just need to specify the neutral element (\code{NilF}) and how to combine two elements (\code{ConsF x y}). All the inner mechanics of how the whole list is reduced, is abstracted away in the \code{cata} function. 
 In the next section we will show different examples of program developed using this pattern.

\subsection{Well-known recursion schemes}

Besides the already mentioned recursion schemes, there are less frequent patterns that hold some useful properties when building recursive programs. The most well-known recursion schemes (including the ones already mentioned) are:

\begin{itemize}
    \item \textbf{catamorphism / anamorphism:} also known as folding and unfolding, respectively. The catamorphism aggregates the information stored in the inductive type. The anamorphism generates an inductive type starting from a seed value.
    \item \textbf{paramorphism / apomorphism:} these Recursion Schemes work as catamorphism and anamorphism, however, at every step they allow access to the original downwards structure. 
    \item \textbf{histomorphism / futumorphism:} histomorphism allows access to every previously consumed elements from the most to the least recent and futumorphism allows access to the elements that are yet to be generated.
\end{itemize}

And, of course, we can also combine these morphisms creating the \emph{hylomorphism} (anamorphism followed by catamorphism), \emph{metamorphism} (catamorphism followed by anamorphism), and chronomorphism (combination of futumorphism and histomorphism).

In the following section we will explain some possible ideas on how to exploit these patterns in the context of program synthesis.

\section{Origami}\label{sec:origami}

The main idea of Origami is is to reduce the search space by breaking down the synthesis process into smaller steps. An overview of the approach is outlined in Algorithm~\ref{alg:origami}.

Step~\textcircled{\raisebox{-0.9pt}1} of the algorithm is to determine the recursion scheme of the program. This can be done heuristically (\emph{e.g.}, following a flowchart\footnote{\url{https://hackage.haskell.org/package/recursion-schemes-5.2.2.4/docs/docs/flowchart.svg}}). Since there are just a few known morphisms and the distribution of use cases for each morphism is highly skewed, this determination could be run in parallel; interactively determined by an expert; chosen by a machine learning algorithm based on the input and output types; derived by the input/output examples; or obtained from the textual description of the algorithm.

Then, in Step~\textcircled{\raisebox{-0.9pt}2} after the choice of recursion scheme, it is time to choose the appropriate base (inductive) data type. The most common choices are natural numbers, lists, and rose trees. Besides these choices, one can provide custom data structures if needed. This choice could be done employing the same same methods used in Step~\textcircled{\raisebox{-0.9pt}1}.

\begin{algorithm}[t!]
\caption{Origami Program Synthesis}\label{alg:origami}
\begin{algorithmic}[1]
\Procedure{Origami}{$x, y, types$}\Comment{The training data and the program type.}
   \State $r\gets \operatorname{pickRecursionScheme(types)}$ \Comment{Step~\textcircled{\raisebox{-0.9pt}1}}
   \State $b\gets \operatorname{pickInductiveType(types)}$ \Comment{Step~\textcircled{\raisebox{-0.9pt}2}}
   \State $p\gets \operatorname{pickTemplate(r, b, types)}$ \Comment{Step~\textcircled{\raisebox{-0.9pt}3}}
   \State $f\gets \operatorname{createFitnessFunction(p, b)}$  \Comment{Step~\textcircled{\raisebox{-0.9pt}4}}
   \State \Return $\operatorname{evolveProgram(p, f)}$ \Comment{Step~\textcircled{\raisebox{-0.9pt}5}}
\EndProcedure
\end{algorithmic}
\end{algorithm}

Step~\textcircled{\raisebox{-0.9pt}3} deals with the choice of which specific template of evolvable functions (further explained in the next sections) will specify the parts of the program that must be evolved returning a template function to be filled by the evolutionary process. Once this is done, we can build the fitness function (Step~\textcircled{\raisebox{-0.9pt}4}) that will receive the evolved functions, wrap them into the recursion scheme, and evaluate them using the training data. Finally, we run the evolution (Step~\textcircled{\raisebox{-0.9pt}5}) to find the correct program.

To illustrate the process, let us go through the process to generate a solution to the problem \emph{count odds} from the General Program Synthesis Benchmark (GPSB)~\cite{helmuth2015general}:


\begin{displayquote}
\textbf{Count Odds} Given a vector of integers, return the number of integers that are odd, without use of a specific even or odd instruction (but allowing instructions such as modulo and quotient).
\end{displayquote}

We can start by determining the type signature of this function:

\begin{lstlisting}
    countOdds :: [Int] -> Int
\end{lstlisting}

\begin{enumerate}
    \item[Step~\textcircled{\raisebox{-0.9pt}1}] As the type signature suggests, we are collapsing a list of values into a value of the same type. So, we should pick one of the \emph{catamorphism} variants. Let us pick the plain catamorphism.
    \item[Step~\textcircled{\raisebox{-0.9pt}2}] In this step we need to choose a base inductive type. Since the type information tells us we are working with lists, we can use the list functor.
    \item[Step~\textcircled{\raisebox{-0.9pt}3}] As the specific template we choose the reduction to a value (from a list of integers to a single value).
\end{enumerate}

We are now at this point of the code generation where we depart from the following template:

\begin{lstlisting}[escapechar=@,basicstyle=\ttfamily]
    countOdds :: [Int] -> Int
    countOdds ys = cata alg (fromList ys)

    alg :: ListF Int Int -> Int
    alg xs = case xs of
               NilF -> @\evolv{e1}@
               ConsF x y -> @\evolv{e2}@
\end{lstlisting}

\begin{enumerate}
    \item[Step~\textcircled{\raisebox{-0.9pt}4}] We still need to fill up the \emph{gaps} \code{e1} and \code{e2} in the code. At this point, the piece of code \code{NilF -> e1} can only evolve to a constant value as it must return an integer and it does not have any integer available in scope. The piece of code \code{ConsF x y -> e2} can only evolve to operations that involve \code{x}, \code{y}, and integer constants.
    \item[Step~\textcircled{\raisebox{-0.9pt}5}] Finally, the evolution can be run and the final solution should be:

\begin{lstlisting}
    countOdds :: [Int] -> Int
    countOdds xs = cata alg xs

    alg :: ListF Int Int -> Int
    alg xs = case xs of
               NilF -> 0
               ConsF x y -> mod x 2 + y
\end{lstlisting}
\end{enumerate}

As we can see from this example, the evolution of the functions inside the recursion scheme is well determined by the input-output types of the main function. We should notice, though, that this translation is not always straightforward. To identify the different templates that can appear, we manually solved the entire GPSB benchmark in such a way that the evolvable \emph{gaps} of the recursion schemes have a well determined and concrete function type. Particularly to this work, we will highlight one example of each template, but the entire set of solutions is available at the Github repository at \url{https://github.com/folivetti/origami-programming}. This repository is in continuous development and will also contain alternative solutions using different data structures (\emph{e.g.}, indexed lists) and solutions to other benchmarks.
In the following examples, the evolvable parts of the solution are shown in underlined green, making it more evident the number and size of programs that must be evolved by the main algorithm.

It should be noted that we made some concessions in the way some programs were solved. In particular, our solutions are only concerned with returning the required values and disregards any IO operations (for instance,  \emph{print the result with a string "The results is"}) as we do not see the relevancy in evolving this part of the program at this point. This will be part of the full algorithm for a fair comparison with the current state-of-the-art.

\subsection{How to choose a template}

The first step requires making a decision among one of the available templates. Some options (from the most naive to more advanced ones) are listed below:

\begin{itemize}
    \item Run multiple searches in parallel with each one of the templates
    \item Integrate this decision as part of the search (\emph{e.g.}, encode into the chromosome)
    \item Use the type information to pre-select a subset of the templates (see. Table~\ref{tab:type-to-rec})
    \item Use the description of the problem together with a language model
\end{itemize}

\begin{table}[t!]
    \centering
        \caption{Association between type signatures and its corresponding recursion schemes.}
    \begin{tabular}{c|c}
       \toprule
       \textbf{Type signature}  & \textbf{Recursion Scheme} \\
       \midrule
        \code{f a -> b} & catamorphism, accumorphism \\
        \code{a -> f b} & anamorphism \\
        \code{a -> b} & hylomorphism \\
        \bottomrule
    \end{tabular}
    \label{tab:type-to-rec}
\end{table}

Specifically to the use of type information, as we can see in Table~\ref{tab:type-to-rec}, the type signature can constrain the possible recursion schemes, thus reducing the search space of this choice. There are also some specific patterns in the description of the program that can help us choose one of the templates. For example, whenever the problem requires returning the position of an element, we should use accumorphism.






\subsection{Jokers to the right: catamorphism}

In the previous sections we gave the definition of a catamorphism. Indeed, its definition is analogue to a right fold but generalized to any fixed point data structure. For a list $x$ with $n$ elements, this is equivalent to applying a function in the following order: $f(x_0, f(x_1, f(x_2, \ldots f(x_{n - 1}, y)\ldots )$, where $y$ is the initial value. So the accumulation is performed from the end of the structure to the beginning. Notice that all of the solutions follow the same main form \code{cata alg (fromList data)}, that changes the input argument into a fixed form of a list and apply the algebra of the catamorphism.
Specifically for the catamorphism, we observed four different templates that we will exemplify in the following, from the simplest to the more complicated approaches. In what follows, we will only present the definition of the \code{alg} function.

\subsubsection{Reducing a structure}

The most common use case of catamorphism is to reduce a structure to a single value, or \code{f a -> a}. In this case the algebra follows a simple function that is applied to each element and combined with the accumulated value. This template was already illustrated in the beginning of Section~\ref{sec:origami} with the example of \emph{countOdds}. Due to space constraint we will refer the reader to that particular example.

\subsubsection{Regenerating the structure: mapping}

The higher-order function \code{map} is a \code{fold} that process and reassembles the structure, so, any function with a type signature \code{f a -> f b} is a catamorphism.

\begin{displayquote}
\textbf{Double Letters} Given a string, print (in our case, return) the string, doubling every letter character, and tripling every exclamation point. All other non-alphabetic and non-exclamation characters should be printed a single time each.
\end{displayquote}

\begin{lstlisting}
    -- required primitives: if-then-else, (<>), ([])
    -- user provided: constant '!', constant "!!!", isLetter
    alg NilF         = @\evolv{[]}@
    alg (ConsF x xs) = @\evolv{if x == '!'}@
                         @\evolv{then "!!!" <> xs}@
                         @\evolv{else if isLetter x then [x,x] <> xs}@
                                         @\evolv{else x:xs}@
\end{lstlisting}

The main difference from the previous example is that, in this program, at every step the intermediate result (\code{xs}) is a list. Notice that the \code{ConsF} case is still constrained in such a way that we can either insert the character \code{x} somewhere in \code{xs}, or change \code{x} into a string and concatenate to the result.

\textbf{Evolvable functions:} given a function of type \code{f a -> f b}, we need to evolve i) the pattern \code{alg NilF} of type \code{f b} and the pattern \code{alg (ConsF x xs)} of type \code{a -> f b -> b}.

\subsubsection{Generating a function}

Another catamorphism template are functions that return a function, or in Haskell notation \code{f a -> f a -> b}, which is read as: a function that takes two arguments of type \code{f a} (\emph{e.g.}, a list of values) and returns a value of type \code{b}. This signature is equivalent to its curried form which is \code{f a -> (f a -> b)}: a function that takes a value of type \code{f a} and returns a function that takes an \code{f a} and returns a value of \code{b}. While generating a function that returns a function seems to add complexity, the type constraints can help guiding the synthesis more efficiently than if we were to interpret it as a function of two arguments.

\begin{displayquote}
\textbf{Super Anagrams} Given strings x and y of lowercase letters, return true if y is a super anagram of x, which is the case if every character in x is in y. To be true, y may contain extra characters, but must have at least as many copies of each character as x does.
\end{displayquote}

\begin{lstlisting}
    -- required primitives: delete, constant bool
    -- elem, (&&)
    alg NilF ys = @\evolv{True}@
    alg (ConsF x xs) ys = @\evolv{(not.null) ys \&\& elem x ys \&\& xs (delete x ys)}@
\end{lstlisting}

For this problem, we incorporated the second argument of the function as a second argument of \code{alg}. For the base case, the end of the first string, we assume that this is a super anagram returning \code{True}. For the second pattern, we must remember that \code{xs} is supposed to be a function that receives a list and return a boolean value. So, we first check that the second argument is not null, that \code{x} is contained in \code{ys} and then evaluate \code{xs} passing \code{ys} after removing the first occurrence of \code{x}.
Notice that for the \code{NilF} case we are not limited to returning a constant value, we can apply any function to the second argument that returns a boolean. Thus, any function \code{String -> Bool} will work. Even though we have more possibilities for the base case, we can grow the tree carefully to achieve a proper solution. The same goes for the second case in which we add more possible programs as we have in  our possession a char value, a string and a function that process a string.

\textbf{Evolvable functions:} given a function of type \code{f a -> f a -> b}, we need to evolve i) the pattern \code{alg NilF} of type \code{f a -> b} and, ii) the pattern \code{alg (ConsF x xs)} of type \code{a -> (f a -> b) -> f a -> b}.

\subsubsection{Combination of patterns}

More complex programs often combine two or more different tasks represented as functions that return tuples (\code{f a -> (b, c)}). If both tasks are independent and they are both catamorphisms, they are equivalent to applying different functions in every element of the $n$-tuple. The evolution process would be the same as per the previous template but we would evolve one function for each output type.

\subsection{When you started off with nothing: anamorphism}

The anamorphism starts from a seed value and then unfolds it into the desired recursive structure. Its evolutionary template is composed of a function over its argument that spans into multiple cases, each one responsible to evolve one function. The most common case is that this main function is a predicate that spans over a \code{True} or \code{False} response. The result of each case should be the return data structure containing one element and one seed value.
For this recursion scheme we only identified a single template in which the first argument is used as the initial seed and any remaining argument (if of the same type) is used as a constant when building the program. Of course, during the program synthesis, we may test any permutation of the use of the input arguments.

\begin{displayquote}
\textbf{For Loop Index} Given 3 integer inputs start, end, and step, print the integers in the sequence $n_0$ = start, $n_i$ = $n_i-1$ + step for each $n_i$ < end, each on their own line.
\end{displayquote}

\begin{lstlisting}
    -- required primitives: (==), (+)
    forLoopIndex :: Int -> Int -> Int -> [Int]
    forLoopIndex start end step = toList (ana coalg @\evolv{start}@)
      where
        coalg seed = case @\evolv{seed == end}@ of
                       True -> NilF
                       False -> ConsF @\evolv{seed (seed + step)}@
\end{lstlisting}

In this program, the first argument is the starting seed of the anamorphism and the step and end is used when defining each case. The \code{case} predicate must evolve a function that takes the seed as an argument and returns a boolean. To evolve such a function we are limited to the logical and comparison operators. As the type of the seed is well determined, we must compare it with values of the same type, which can be constants or one of the remaining arguments. After that, we must evolve two programs, one that creates the element out of the seed (a function of type \code{Int -> Int}) and the generation of the next seed.

\textbf{Evolvable functions:} given a function of type \code{a -> a -> ... -> a -> f b}, we need to evolve i) the pattern \code{coalg} of type \code{a -> f b a}, when using list as the base type, ii) the predicate function \code{a -> Bool} that returns \code{True} for the terminating case (if any), iii) the starting seed value (either a constant or one of the arguments).

\subsection{Stuck in the middle with you: hylomorphism}

Hylomorphism is the fusion of both catamorphism and anamorphism. This template works the same as evolving the functions for both schemes.

\begin{displayquote}
\textbf{Collatz Numbers} Given an integer, find the number of terms in the Collatz (hailstone) sequence starting from that integer.
\end{displayquote}

\begin{lstlisting}
  -- required primitives: constant int, (==)
  -- (+), (*), mod, div
  alg NilF = @\evolv{1}@
  alg (ConsF x xs) = @\evolv{1 + xs}@

  coalg x =
    case @\evolv{x == 1}@ of
      True -> NilF
      False -> ConsF x @\evolv{(if mod x 2 == 0}@
                          @\evolv{then div x 2 }@
                          @\evolv{else div (3*x + 1) 2)}@
\end{lstlisting}

In hylomorphism, first the coalgebra produces a value that can be consumed by the algebra. The the single input argument will be the seed to generate the next hailstone number as the next seed. If this seed is equal to $1$ the process terminates. The algebra in this case simply counts the number of generated values but adding a $+1$ (\code{NilF}) to account for the value $1$ that was dropped during the anamorphism.

\textbf{Evolvable functions:} given a function of type \code{a -> b}, we need to evolve i) the pattern \code{coalg} of type \code{a -> f a b}, ii) the pattern \code{alg (ConsF x xs)} of type \code{f a b -> b}, and iii) the pattern \code{alg NilF} of type \code{b}.

\subsection{Clowns to the left of me: accumorphism}

In some situations our solution needs to traverse the inductive structure to the left. In other words, at every step we accumulate the results and gain access to the partial results. For lists, this is equivalent to $f(\ldots f(f(y, x_0), x_1), \ldots x_n)$.
For this purpose we can implement \emph{accumorphism} that requires an algebra and an accumulator.
This recursion scheme requires an accumulator function \code{st} besides the algebra. Notice that the accumulator function will receive a fixed form structure, the current state, and it will return the fixed form with a tuple of the original value and the trace state. 
This template must be carefully used because it can add an additional degree of freedom through the \code{st} function, notice that this function can be of any type, not limited by any of the main program types, thus expanding the search space. To avoid such problem, we will use accumulators in very specific use-cases as described in the following sub-sections.

\subsubsection{Indexing data}

Whenever the problem requires the indexing of the data structure, we can use the accumulator to store the index of each value of the structure and, afterwards, use this information to process the final solution. With this template, the accumulator should be of type \code{Int}.

\begin{displayquote}
\textbf{Last Index of Zero} Given a vector of integers, at least one of which is 0, return the index of the last occurrence of 0 in the vector.
\end{displayquote}

\begin{lstlisting}
    -- required primitives: if-then-else, (+), (==)
    -- (<>), constant int, Maybe, Last
    lastIndexZero :: [Int] -> Int
    lastIndexZero xs = accu st alg (fromList xs) @\evolv{0}@
      where
        st NilF s = NilF
        st (ConsF x xs) s = ConsF x (xs, @\evolv{s+1}@)

        alg NilF s = @\evolv{-1}@
        alg (ConsF x xs) s = @\evolv{if x == 0 \&\& xs == -1}@
                               @\evolv{then s}@
                               @\evolv{else xs}@
\end{lstlisting}

This template requires that we evolve: the initial value for \code{s}, the expression to update \code{s} at every element of the list (the remainder of \code{st} function is given), the value of the base case of \code{alg}, and the general case.
The accumulator function of this program has the function of indexing our list. When the list is indexed, we build the result from bottom up by signaling that we have not found a zero by initially returning \code{-1}. Whenever \code{x == 0} and the current stored index is \code{-1}, the program returns the index stored in that level (\code{s}). Otherwise, it just returns the current  \code{xs}. 

\textbf{Evolvable functions:} given a function of type \code{f a -> b}, we need to evolve i) a constant value of type \code{Int}, ii) the pattern \code{st (ConsF x xs) s} of type \code{a -> Int -> (a, Int)}, iii) the pattern \code{alg NilF s} of type \code{Int -> b}, iv) the pattern \code{alg (ConsF x xs) s} of type \code{a -> b -> Int -> b}.

\subsubsection{A combination of catamorphisms}

In some cases the recursive function is equivalent to the processing of two or more catamorphisms, with a post-processing step that combines the results. A simple example is the average of the values of a vector in which we need to sum the values and count the length of the vector, combining both final results with the division operator. This template of catamorphism constrains the type of the accumulator to a tuple of the returning type of the program.

\begin{displayquote}
\textbf{Vector Average} Given a vector of floats, return the average of those folats. Results are rounded to 4 decimal places.
\end{displayquote}

\begin{lstlisting}
    -- required primitives: (+), (/)
    vecAvg :: [Double] -> Double
    vecAvg xs = accu st alg (fromList xs) @\evolv{(0.0, 0.0)}@
      where
        st NilF (s1, s2) = NilF
        st (ConsF x xs)  = ConsF x (xs, (@\evolv{s1 + x}@, @\evolv{s2 + 1)}@)

        alg NilF (s1, s2) = @\evolv{s1 / s2}@
        alg (ConsF x xs) s = xs
\end{lstlisting}
We, again, illustrate this solution by splitting the accumulator function into two distinct functions, one for each element of the tuple. While function \code{f} accumulates the sum of the values of the list, function \code{g} increments the accumulator by one at every step. In this template, the final solution is the combination of the values at the final state of the accumulator, thus in the \code{alg} function we just need to evolve a function of the elements of the state.

\textbf{Evolvable functions:} given a function of type \code{f a -> b}, we need to evolve i) a constant value of type \code{(b, b)}, ii) the pattern \code{st (ConsF x xs) s} of type \code{a -> (b,b) -> (b, b)}, iii) the pattern \code{alg NilF s} of type \code{(b,b) -> b}.

In the next section we report a simple experiment with a subset of these problems as a proof-of-concept of our approach.

\vspace{-1em}\section{Preliminary Results}\label{sec:preliminary}

The main objective of this work is to introduce the ideas of using recursion schemes to solve programming challenges and to verify whether the current benchmark problems can be solved using this approach. 
In this section, we will show how using the catamorphism template can help improve the overall performance of a GP approach. For this purpose, we will use the Higher-Order Typed Genetic Programming (HOTGP)~\cite{fernandes2023hotgp}. HOTGP supports higher-order functions and $\lambda$-functions, disallows the creation of impure functions, and uses type information to guide the search.


We adapted this algorithm to generate only the \emph{evolvable} parts of a catamorphism (in here, implemented as a \code{foldr}) and tested most of the benchmarks that can be solved by this specific template. 
Specifically, we asked HOTGP to generate an expression for the \code{alg (ConsF x xs)} pattern. For the \code{alg NilF}, we used default \emph{empty} values depending on the data-type: $0$ for \code{Int} and \code{Float}, \code{False} for \code{Bool}, the space character for \code{Char}, and empty lists for lists and strings. Naturally, this is a simplification of the template, as all of the benchmarks we are interested in happen to use these values for the null pattern. In order to properly generate the recursion patterns, this part of the function should also be considered in the evolution.

We set the maximum depth of the tree to 5, as the expressions we want to generate are always smaller than that. All the other parameters use the same values described by Fernandes et al.~\cite{fernandes2023hotgp}.
To position Origami within the current literature, we compare the obtained results against those obtained by HOTGP itself~\cite{fernandes2023hotgp}, PushGP in the original benchmark~\cite{helmuth2015detailed}, Grammar-Guided Genetic Programming (G3P)~\cite{g3p}, and the extended grammar version of G3P (here called G3P+)~\cite{g3pe}, as well as some recently proposed methods such as Code Building Genetic Programming (CBGP)~\cite{pantridge2022functional}, G3P with Haskell and Python grammars (G3Phs and G3Ppy)~\cite{garrow2022functional}, Down-sampled lexicase selection (DSLS)~\cite{dsls} and Uniform Mutation by Addition and Deletion (UMAD)~\cite{umad}.

We removed the benchmarks that would need the algorithm to output a function (Mirror Image, Vectors Summed and Grade), as this is not currently supported by HOTGP.
For completeness, we also tested the benchmarks that can be solved by accumorphisms, even though this adaptation does not support it to show that once committed to a template (\emph{e.g.}, catamorphism), the algorithm cannot find the correct solution if it requires a different template (\emph{e.g.}, accumorphism).

Analyzing the results depicted in Table~\ref{tab:results}, one can notice that when comparing the standard HOTGP with Origami, Origami always obtains an equal or better number of perfect runs, except on the accumorphism benchmarks. Not only that, but the number of problems that are always solved increased from $1$ to $4$, and those higher than $75\%$ increased from $3$ to $6$, a significant improvement in success rate. When compared to related work, out of the $7$ solvable benchmarks, Origami had the best results in $6$ of them. The only exception being the \emph{replace-space-with-newline}. Overall, once we choose the correct template, the synthesis step becomes simpler.


\begin{table}[t]
    \centering
    \caption{Percentage of runs that returned a perfect solution on the validation set. The bottom part of the table summarizes the result as the number of times each algorithm had the highest percentage, and in how many problems the percentage was greater or equal to a certain threshold. The benchmarks marked with $*$ are only solvable with accumorphism.}
\resizebox{0.9\textwidth}{!}{\begin{minipage}{\textwidth}
    \hspace{-4.5mm}\begin{tabular}{lrrrrrrrrrr}
    \toprule
    Benchmark & Origami & HOTGP & DSLS & UMAD & PushGP & G3P & CBGP & G3P+ & G3Phs & G3Ppy \\
    \midrule
    checksum*& 0 & -- & 1 & \underline{5} & 0 & 0 & -- & 0 & -- & -- \\
    count-odds & \underline{100} & 50 & 11 & 12 & 8 & 12 & 0 & 3 & -- & -- \\
    double-letters & \underline{94} & 0 & 50 & 20 & 6 & 0 & -- & 0 & -- & -- \\
    last-index-of-zero* & 0 & 0 & \underline{62} & 56 & 21 & 22 & 10 & 44 & 0 & 2 \\
    negative-to-zero & \underline{100} & \underline{100} & 82 & 82 & 45 & 63 & 99 & 13 & 0 & 66 \\
    replace-space-with-newline & 60 & 38 & \underline{100} & 87 & 51 & 0 & 0 & 16 & -- & -- \\
    scrabble-score & \underline{100} & -- & 31 & 20 & 2 & 2 & -- & 1 & -- & -- \\
    string-lengths-backwards & \underline{100} & 89 & 95 & 86 & 66 & 68 & -- & 18 & 0 & 34 \\
    syllables & \underline{84} & 0 & 64 & 48 & 18 & 0 & -- & 39 & -- & -- \\
    vector-average* & 0 & 80 & 88 & \underline{92} & 16 & 5 & 88 & 0 & 4 & 0 \\
\midrule
\textbf{\# of Best Results} & 6 & 1 & 2 & 2 & 0 & 0 & 0 & 0 & 0 \\
\midrule
    $\mathbf{=100}$ & \underline{4} & 1 & 1 & 0 & 0 & 0 & 0 & 0 & 0 & 0 \\
    $\mathbf{\geq75}$ & \underline{6} & 3 & 4 & 4 & 0 & 0 & 2 & 0 & 0 & 0 \\
    $\mathbf{\geq50}$ & \underline{7} & 4 & \underline{7} & 5 & 2 & 2 & 2 & 0 & 0 & 1 \\
    $\mathbf{>0}$ & 7 & 5 & \underline{10} & \underline{10} & 9 & 6 & 3 & 7 & 1 & 3 \\
    \bottomrule
    \end{tabular}
\end{minipage}}    
    \label{tab:results}
\end{table}

\vspace{-1em}\section{Discussion and Final Remarks}\label{sec:discussion}

Our main hypothesis with this study is that, by starting the program synthesis fixing one of the recursion schemes, we simplify the process of program synthesis. For this purpose we used a general set of benchmarks widely used in the literature. Within this benchmark suite, we observed that, in most cases, the evolvable part of the programs becomes much simpler, to the point of being trivial. However, some of them require a pre-processing of the input arguments with some general use functions (such as \code{zip}) to keep this simplicity, or an adaptation to the output type as to return a function instead of a value. 

In some cases, more complicated functions can be evolved with the help of a human interaction by asking additional information such as \emph{when should the recursion stop?}. Also, in many cases, the type signature of each one of the evolvable programs already constrains the search space. For example, the pattern \code{alg NilF} must return a value of the return type of the program without using any additional information, thus the space is constrained to constant values of the return type.

Analysing the minimal function set required to solve all these problems, one can formulate a basic idea about the adequate choice based on the signature of the main function and on any user-provided type/function. Table~\ref{tab:functionset} shows the set we used for the presented experimental evaluation.

\begin{table}[t!]
    \centering
    \caption{Function set used for solving the GPSB benchmark problems. }
    \begin{tabular}{c|p{\dimexpr 0.75\linewidth-2\tabcolsep}}
       Type class  & Functions \\
       \hline
        Numbers & \code{fromIntegral, +, -, *, /, \^{}} \code{div, quot, mod, rem} \code{abs, min, max}\\
        Logical & \code{<, <=, >, >=, ==, /=, \&\&, not, ||} \\
        Lists & \code{cons, snoc, <>, head, tail, init, last, null, length, delete, elem}\\
        Tuple & \code{fst, snd} \\
        Map & \code{findMap, insertWith} \\
        General purpose & \code{if-then-else, case, uncurry, fromEnum, toEnum, id} \\
        \hline
    \end{tabular}
    \label{tab:functionset}
    \vspace{-1em}
\end{table}

\begin{table}[b!]
    \centering
    \vspace{-1em}
    \caption{Function set assumed to be provided by the user. All of these functions and constants are explicitly mentioned in the problem description.}
    \begin{tabular}{c|p{\dimexpr 0.75\linewidth-2\tabcolsep}}
       Type  & Functions \\
       \hline
        \code{Int -> Bool} & \code{(< 1000), (>= 2000)}\\
        \code{String} & \code{"small", "large", "!!!", "ABCDF", "ay"} \\
        \code{Char} & \code{'!', ' ', '\textbackslash n'} \\
        \code{Int} & 0, 1, 64 \\
        \code{Char -> Bool} & \code{isVowel, isLetter} \\
        \hline
    \end{tabular}
    \label{tab:userset}
\end{table}

Table~\ref{tab:userset} lists the set of functions and constants we assume that should be provided by the user, as they are contained in the problem description. Some of them can be replaced by \code{case-of} instructions (e.g., \code{isVowel, isLetter, scrabbleScore}), which can increase the difficulty of obtaining a solution. 

\begin{figure}[t!]
    \centering
    \sidecaption[t]
    \includegraphics[width=4.5cm]{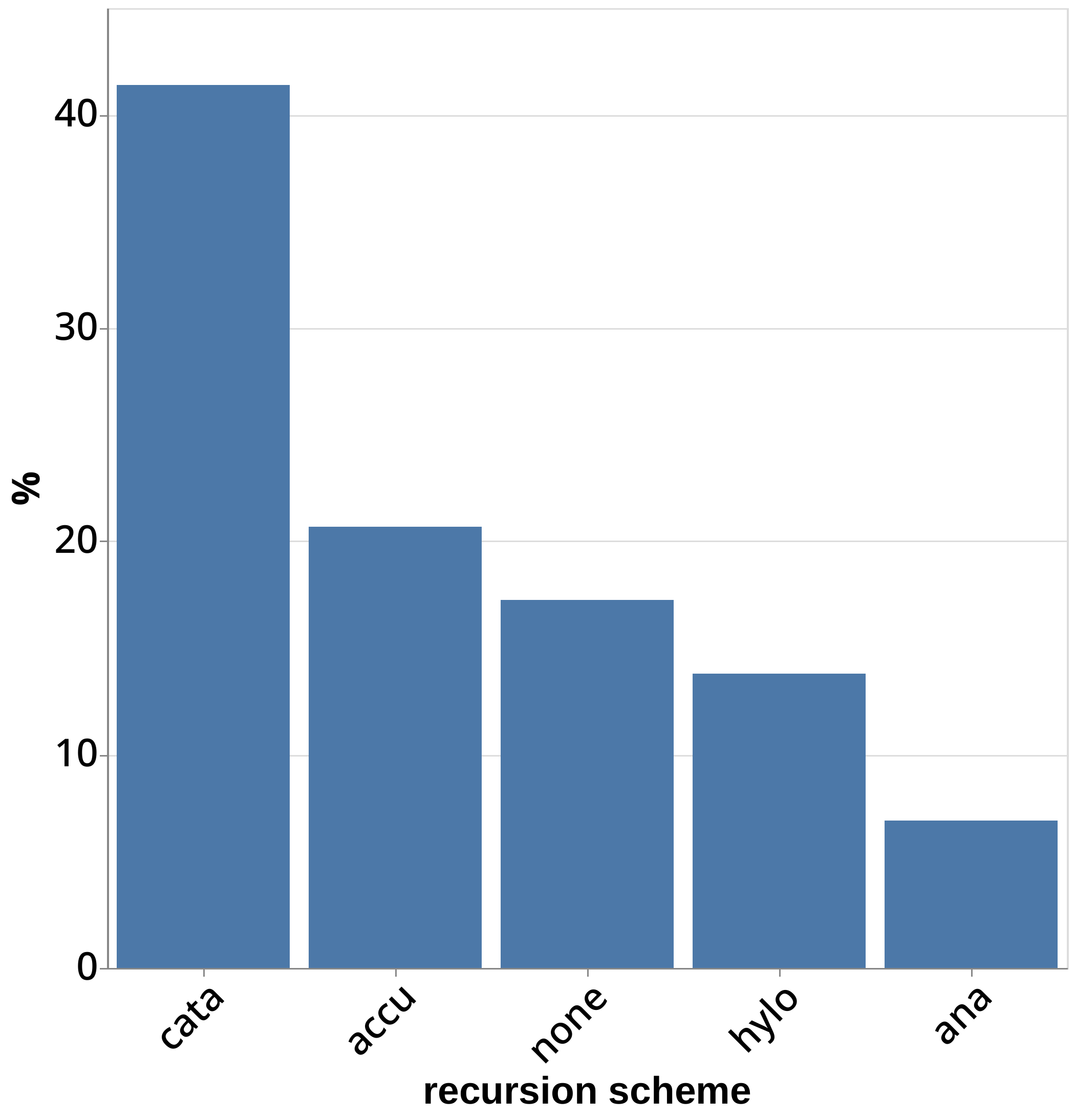}
    \caption{Distribution of recursion schemes used to solve the full set of GPSB problems.}
    \label{fig:dist-morphisms}
\end{figure}

We should note that, from the $29$ problems considered here and implemented by a human programmer, $17\%$ were trivial enough and did not require any recursion scheme; $41\%$ were solved using catamorphism; $20\%$ of used accumorphism (although we were required to constrain the accumulator function). Anamorphism accounted for only $7\%$ of the problems and hylomorphism for $14\%$. The distribution of the usage of each recursion scheme is shown in Figure~\ref{fig:dist-morphisms}. As the problem becomes more difficult and other patterns emerge, we can resort to more advanced recursions such as \emph{dynomorphism} when dealing with dynamic programming problems, for example. Also, none of these problems required a recursive pattern with a base structure different from a list. In the future, we plan to test other benchmarks and introduce new ones that require different structures to test our approach.

One challenge to this approach is how to treat the templates containing multiple evolvable parts. For example, anamorphism requires the evolution of three programs: one that generates the next element, one to generate the next seed, and one predicate to check for the stop condition. We will consider a multi-gene approach~\cite{searson2010gptips} or a collaborative co-evolution strategy~\cite{soule2008improving,grefenstette1996methods}.

As a final consideration, we highlight the fact that most of the programs can be further simplified if we annotate the output type with monoids. In functional programming, and Haskell in particular, monoids are a class of types that have an identity value (\code{mempty}) and a binary operator (\code{<>}) such that \code{mempty <> a = a <> mempty = a}. With these definitions we can replace many of the functions and constants described in Tables~\ref{tab:functionset}~and~\ref{tab:userset} with \code{mempty} and \code{<>}, reducing the search space.

\begin{acknowledgement}
This research was partially funded by Funda\c{c}\~{a}o de Amparo \`{a} Pesquisa do Estado de S\~{a}o Paulo (FAPESP), grant numbers \#2021/12706-1, \#2019/26702-8, \#2021/06867-2, Coordena\c{c}\~{a}o de Aperfei\c{c}oamento de Pessoal de N\'{i}vel Superior - Brasil (CAPES) and CNPq grant number 301596/2022-0.
\end{acknowledgement}
\end{document}